\documentclass{article}

\usepackage[final]{template_custom}
\usepackage[utf8]{inputenc} % allow utf-8 input
\usepackage[T1]{fontenc}    % use 8-bit T1 fonts
\usepackage{hyperref}       % hyperlinks
\usepackage{url}            % simple URL typesetting
\usepackage{booktabs}       % professional-quality tables
\usepackage{amsfonts}       % blackboard math symbols
\usepackage{nicefrac}       % compact symbols for 1/2, etc.
\usepackage{microtype}      % microtypography
\usepackage{xcolor}         % colors

\usepackage{graphicx}
\usepackage{tikz}
\usepackage{comment}
\usepackage{amsmath,amssymb}
\usepackage[accsupp]{axessibility} 
\usepackage{subcaption}
\usepackage{algorithm}
\usepackage{algorithmic}
\usepackage{bm}
\usepackage{tabulary}
\usepackage{colortbl}
\usepackage{amssymb} %checkmark
\usepackage{pifont} % redxmark
\definecolor{darkergreen}{RGB}{21, 152, 56}
\definecolor{red2}{RGB}{252, 54, 65}
\definecolor{Gray}{gray}{0.9}
\definecolor{citecolor}{RGB}{127,255,127}
\definecolor{LightCyan}{rgb}{0.92,1,1}
\newcommand\redp[1]{\textcolor{red2}{(#1)}}
\newcommand\greenp[1]{\textcolor{darkergreen}{(#1)}}

\usepackage{booktabs,tabularx}
\usepackage{siunitx}
\usepackage{xspace}
\newcolumntype{C}{>{\centering\arraybackslash}X}

\def\ModelName{\textsc{TemporalSwap}}
\newcolumntype{x}[1]{>{\centering\arraybackslash}p{#1pt}}
\newlength\savewidth\newcommand\shline{\noalign{\global\savewidth\arrayrulewidth
  \global\arrayrulewidth 1pt}\hline\noalign{\global\arrayrulewidth\savewidth}}

\makeatletter\renewcommand\paragraph{\@startsection{paragraph}{4}{\z@}
  {.5em \@plus1ex \@minus.2ex}{-.5em}{\normalfont\normalsize\bfseries}}\makeatother
 
\newcolumntype{C}[1]{>{\centering\arraybackslash}p{#1}}
\newcolumntype{R}[1]{>{\raggedleft\arraybackslash}p{#1}}
\newcolumntype{L}[1]{>{\raggedright\arraybackslash}p{#1}}
\definecolor{demphcolor}{RGB}{144,144,144}
\newcommand{\demph}[1]{\textcolor{demphcolor}{#1}}

\newcommand*\colorcmark[1]{%
  \expandafter\newcommand\csname #1cmark\endcsname{\textcolor{#1}{\ding{51}}}%
}
\colorcmark{green}

\newcommand*\colorxmark[1]{%
  \expandafter\newcommand\csname #1xmark\endcsname{\textcolor{#1}{\ding{55}}}%
}
\colorxmark{red}
\def \ModelName          {\textit{All-in-one}\xspace}
\def \ModelNameTiny      {\textit{All-in-one}-Ti\xspace}
\def \ModelNameSmall      {\textit{All-in-one}-S\xspace}
\def \ModelNameBase      {\textit{All-in-one}-B\xspace}

\def \ModelNameLarge     {\textit{All-in-one}-L\xspace}
\def \pzo {\phantom{0}} % 
\def \GlobalTableRescale {0.85}

\title{All in One: \\Exploring Unified Video-Language Pre-training}

\author{%
  Alex Jinpeng Wang$^{1}$, Yixiao Ge$^{2}$, Rui Yan$^{1}$, Yuying Ge$^{4}$, Xudong Lin$^{5}$, \\ \textbf{Guanyu Cai$^{1,6}$, Jianping Wu$^{7}$, Ying Shan$^{2}$, Xiaohu Qie$^{3}$, Mike Zheng Shou$^{1}$\thanks{Corresponding Author.}} \\[3pt]
    $^1$Show Lab, National University of Singapore\quad $^2$ARC Lab,$^3$Tencent PCG\quad \\ $^4$The University of Hong Kong\quad $^5$Columbia University\quad $^6$Tongji University\quad $^7$Tsinghua University\\
}

\begin{document}

\maketitle

\begin{abstract}
Mainstream Video-Language Pre-training models \cite{actbert,clipbert,violet} consist of three parts, a video encoder, a text encoder, and a video-text fusion Transformer. They pursue better performance via utilizing heavier unimodal encoders or multimodal fusion Transformers, resulting in increased parameters with lower efficiency in downstream tasks. In this work, we for the first time introduce an end-to-end video-language model, namely \textit{all-in-one Transformer}, that embeds raw video and textual signals into joint representations using a unified backbone architecture. We argue that the unique temporal information of video data turns out to be a key barrier hindering the design of a modality-agnostic Transformer. To overcome the challenge, we introduce a novel and effective token rolling operation to encode temporal representations from video clips in a non-parametric manner. The careful design enables the representation learning of both video-text multimodal inputs and unimodal inputs using a unified backbone model. Our pre-trained all-in-one Transformer is transferred to various downstream video-text tasks after fine-tuning, including text-video retrieval, video-question answering, multiple choice and visual commonsense reasoning. State-of-the-art performances with the minimal model FLOPs on nine datasets demonstrate the superiority of our method compared to the competitive counterparts.
The code and pretrained model have been released in \textcolor{red}{\url{https://github.com/showlab/all-in-one}}.
\end{abstract}

\section{Introduction}

The pre-train-and-then-fine-tune scheme has become a standard paradigm to learn transferable video-language representations for a wide range of downstream video-text tasks, for example, text-video retrieval \cite{msrvtt,msvd}, video-question answering \cite{tgifqa,msrvttqamsvdqa}, multiple choice \cite{tgifqa,msrvttqamsvdqa} and visual commonsense reasoning \cite{merlot}. 
In recent years, there has been tremendous progress in the development of video-language pre-training (VLP) models \cite{clipbert,actbert,actionclip,violet}, where joint representations are generally produced with a multimodal fusion Transformer network after extracting the visual and language features through unimodal encoders.

Mainstream VLP methods attempt to boost the pre-training in two ways: \emph{i}. adopting more expensive video/text encoders to obtain more powerful unimodal features \cite{frozen,violet}
\emph{ii}. designing heavier fusion networks to enhance the associations between modalities \cite{actbert,merlot}.
Despite their advanced performance, they suffer from the increasing parameters, leading to significant computational inefficiency in downstream tasks.

To this end, we aim to \textit{design the simplest and most lightweight video-language model that gathers all capabilities in one}, that is, learning video-language representations from their raw inputs in an end-to-end manner. 
In this way, we do not need any extra unimodal encoders (\textit{e.g.}, object detector in \cite{actbert} or ResNet visual encoder in \cite{clipbert}), and embed visual and text signals in a shared and unified model, termed as \ModelName Transformer in our paper.
A recent study, ViLT \cite{vilt}, accomplishes such end-to-end joint learning in image-text pre-training under the presumption that the Transformer can process images in the same way as it processes text.
However, we observe that it is non-trivial to embed videos using a unified Transformer that is also applied to process textual signals due to the unique challenge of modeling temporal information in video.

\begin{figure}[t]
  \centering
    \includegraphics[width=.95\linewidth]{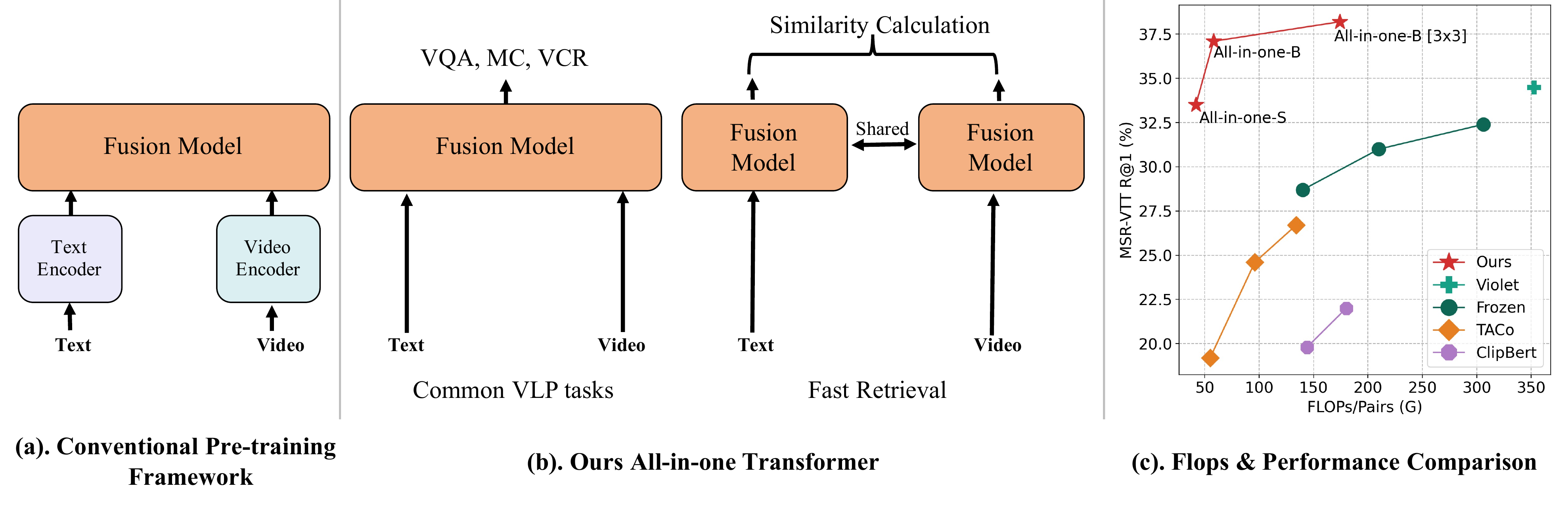}
   \caption{
   \textbf{Compare to mainstream video-language pre-training methods.}
   {(a). Conventional methods \cite{actbert,clipbert,violet,frozen}}
   use deep features from separate encoders before fusion. The fusion layer can be light \cite{frozen} or heavy \cite{actbert,clipbert,violet}.
%   ActBERT \cite{actbert}, ClipBERT \cite{clipbert} and VIOLET \cite{violet}.
   {(b). Ours all-in-one Transformer}
   learns video and text joint representations end-to-end from their raw inputs.
%   one cross-modality transformer merely.
   We also support fast retrieval by feeding unimodal inputs during inference.
%   with shared fusion model.
   (c). Comparison of FLOPs and retrieval performance on MSRVTT \cite{msrvtt}.
   Our \ModelName brings excellent results with modest computational cost.
   }
   \vspace{-1em}
   \label{fig:1_motivation}
\end{figure}

% \begin{figure}[t]
%   \centering
%     \includegraphics[width=.85\linewidth]{figures/src/introduction.pdf}
%   \caption{
%   \textbf{Different ways to conduct video-and-language pre-training: }
%   \textbf{(a). One-stream based approaches}
%   use deep feature from off-the-shelf video encoder like ActBERT \cite{actbert}, ClipBERT \cite{clipbert} and VIOLET \cite{violet}.
%   \textbf{(b). Dual-stream frameworks}
%   contain one text encoder and one video encoder like Frozen \cite{frozen}.
%   \textbf{(c). Ours independent-encoder free one-stream \ModelName} learn one cross-modality transformer merely.
%   The dotted lines means may not contain.
%   }
%   \label{fig:1_motivation}
% \end{figure}

% \begin{figure}[t]
%   \centering
%     \includegraphics[width=.85\linewidth]{figures/src/introduction.pdf}
%   \caption{
%   \textbf{Different ways to conduct cross-modality information fusion: }
%   \textbf{(a). Region Feature based module.}
%   Traditional method use offline object feature like ActBERT \cite{actbert}.
%   \textbf{(b). Grid Feature based module.}
%   One text encoder and one video encoder like ClipBERT \cite{clipbert} and VIOLET \cite{violet}.
%   \textbf{(c). Ours early-fusion \ModelName.}
%   Only one cross-modality transformer is learned.
%   The dotted lines means may not contain.
%   }
%   \label{fig:1_motivation}
% \end{figure}

Existing works encode temporal representations in video-language pre-training via designing temporal attention layers \cite{frozen} or using temporal-aware visual encoders (\textit{e.g.}, 3D convnets in \cite{actbert} or video Transformer in \cite{violet}), which are all infeasible to be applied in our \ModelName Transformer as they are modality-dependent.
To tackle the challenge, \textit{we introduce a novel, effective and flexible method, the temporal token rolling operation, to properly and gradually capture temporal representations in a non-parametric manner}.
Specifically, a proportion of the visual tokens in each frame of a sparsely sampled video clip are cyclic scrolling from frame to frame.
Thus, visual tokens of a certain frame and its corresponding text tokens can ``view'' temporal dynamics from the rolling tokens of other frames through self-attention layers that naturally occur in the Transformer architecture.
A sub-optimal solution to this issue is to aggregate all frames' visual tokens for the self-attention layer, which is inflexible as it increases the time complexity by a factor of $k$ (given $k$ frames per video clip) compared to our method.

Our \ModelName Transformer is pre-trained towards the objectives of video-text matching and masked language modeling, following the common practice of Image-Language models \cite{vilt}.
The pre-trained model is then fine-tuned to perform downstream video-text tasks. 
To further reap the modality-agnostic benefit of our \ModelName Transformer, \textit{we claim that our pre-trained model can not only encode the joint representations of video-language multimodal inputs, but also embed the unimodal features by feeding only video or text data into the Transformer}.
By fine-tuning our pre-trained model with a contrastive loss between video and text features, our \ModelName Transformer can play the role of an ordinary dual-stream framework \cite{frozen} on the downstream text-video retrieval tasks, realizing fast retrieval.

Our contributions are summarized as three-fold.
(1) We introduce the simplest, most lightweight, and most efficient video-language model for pre-training, namely \ModelName Transformer, which is the first to capture video-language representations from the raw visual and textual signals end-to-end in a unified backbone architecture.
(2) We elucidate and tackle the difficulties of applying a unified and shared backbone for multimodal video and text data, that is, how to properly process the unique temporal information of videos. A novel temporal token rolling operation is proposed to capture the temporal representations of sparsely sampled frames without any extra parameters or increasing time complexity.
(3) Comprehensive experiments on four downstream video-text tasks of nine datasets fully demonstrate the superiority of our pre-trained \ModelName Transformer on both effectiveness and efficiency compared to recent mainstream methods \cite{clipbert,frozen}. Moreover, benefiting from the modality-agnostic characteristic of our model, our pre-trained Transformer can be treated as a dual-stream framework to encode separate video and text features for highly efficient retrieval.

\section{Related Work}

\noindent\textbf{Video-Language Pre-training.}
Pre-training on large-scale video-text pairs and fine-tuning on specific downstream tasks gradually becomes the standard paradigm in the video-language domain.
Pre-trained models show strong transfer ability in a series of popular downstream video-language tasks including Text-to-Video Retrieval~\cite{msrvtt,msvd,bridgeformer,regionlearner}, Video Question Answering~\cite{tgifqa,msrvttqamsvdqa}, and Visual Storytelling~\cite{merlot}.
Previous approaches~\cite{actbert,sun2019videobert,taco} leverage offline video and text features extracted from off-the-shelf visual and language backbones. 
Some recent methods including ClipBERT~\cite{clipbert} and Frozen~\cite{frozen} have attempted to train models in an end-to-end fashion but still rely on well-trained visual encoders for feature extraction.
In addition, these works mainly pre-train models on image-text dataset, like Google Conceptual Captions \cite{cc3m} and Visual genome \cite{visualgenome}, and finetune the pre-trained models for downstream video-language tasks.
In this work, we try to challenge this paradigm and focus on exploring effective strategies for pre-training on pure large-scale video-text benchmarks with only one network, and adapt our approach to various video-language downstream tasks.

\noindent\textbf{Temporal Modeling in Video Understanding.}
Temporal modeling is a fundamental yet challenging topic in video representation learning. 
Several classic ideas including sparse sampling~\cite{tsn}, 3D-type operations~\cite{i3d,videoswin} are proposed for temporal modeling in both convolution and Transformer architectures. 
3D-type temporal modeling like Timesformer~\cite{timesformer} is extremely time-consuming because of the increasing number of sampled frames, which can be disastrous for large-scale pre-training techniques.
Sparse sampling along the temporal dimension, a type of data augmentation proposed in TSN~\cite{tsn}, has been widely adopted to train video backbones. 
Based on this, more related works \cite{tsm,actionclip} try to shift channels among different frames for temporal modeling in action recognition.
Inspired by these works, we try to roll video tokens for better alignment between modalities.
This work focuses on parameter-free temporal modeling based on sparsely sampled frames without heavy 3D-type operation.

\noindent\textbf{Unified Architecture Design for Multimodal Data.}
Recently the unified model, which is capable of processing either unimodal inputs or multimodal inputs with a shared encoder, has attracted a lot of attention.
VATT \cite{vatt} trains a shared transformer with unimodal inputs to process Video, Audio and Text via multimodal contrastive learning and improves the performance of action recognition.
Omnivore \cite{omnivore} converts image, video, and single-view 3D modalities into embeddings that are fed into a Transformer model and trains the model with multi-task learning, which focuses on image/video/scene classification.
In image-text pre-training, the early work Unimo \cite{unimo} solves both understanding and generation tasks with cross-modal contrastive learning.
More recently, UFO \cite{ufo} also uses contrastive learning and employs a momentum teacher to guide the pre-training of a image-text shared encoder, which incurs large computational costs.
Based on cross-modal contrastive learning, our work can also process unimodal inputs and perform retrieval tasks in a dual stream manner, which is very efficient.
To the best of our knowledge, \ModelName Transformer is the first unified network for video-language pre-training.

\section{Method}

\begin{figure}[t]
  \centering
    \includegraphics[width=.9\linewidth]{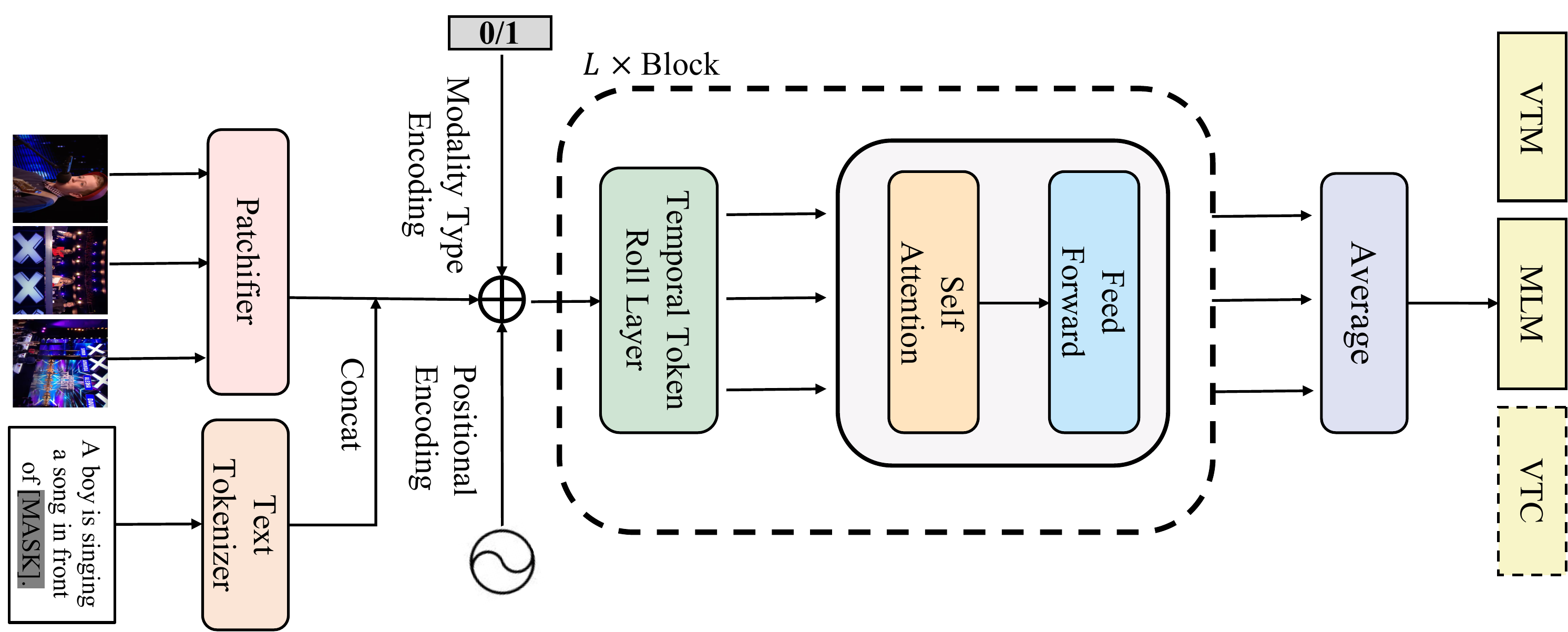}
  \caption{
  \textbf{Model overview.} 
  The overall framework is based on ViT \cite{vit} and only the light text tokenizer and the task head adds extra parameters. 
  The temporal token rolling layer is introduced before each self-attention block to model temporal information. For simplicity, the normalization layers are omitted. 
  %By introducing Temporal Token Roll Layer, the model not only learns correspondence intra-frame (in \textcolor{cyan}{cyan} color) but also inter-frame (in \textcolor{red}{red} color) with modest computation cost. 
  % \xl{maybe change 12 to M to be generic}
  }
%  \vspace{-2em}
  \label{fig:3_ppl}
\end{figure}

We propose \ModelName Transformer, a generic framework that enables end-to-end learning on video and language data, by learning joint representations directly from raw video pixels and raw text tokens, instead of the deeper feature from separate deep embedder.
\ModelName has a succinct architecture as a Video-Language Pre-training model with parameter-free temporal modeling layer.
In model design we making the pipeline as simple as possible so that the model can be used almost out of the box.

\subsection{Unified Video-language Transformer}
Fig.\ref{fig:3_ppl} gives an overview of \ModelName framework. 
It adopts a \textit{sparse sampling strategy} using only $S$ segments (one frame in each segment) at each training step, instead of full-length videos. 
Formally, we denote a video-text pair as $v \in \mathbf{R}^{S \times C\times H \times W}$ (for video) and $t \in \mathbf{R}^{P\times|V|}$ (for text sequence), where $C$ is the number of channels, $(H,W)$ is the resolution of each raw frame, $P$ is the length of input sentence and $|V|$ is the length of the word dictionary.

The Transformer uses constant latent vector size $D$ through all of its layers, so we map both video and text to $D$ dimensions.
Specifically, the input video $v$ is sliced into patches and flatten to $v \in \mathbb{R}^{N\times (P^2C)}$, where $(P,P)$ is the patch resolution and $N=SHW/P^2$.
Followed by linear projection $V \in \mathbb{R}^{(P^2C) \times D}$ for video patches.
And text $t$ with a learned word embedding matrix $T \in \mathbb{R}^{|V|\times D}$
In this way, $v$ is embedded into $\hat{v} \in \mathbb{R}^{N\times D}$ and $t$ is embedded into $\hat{t} \in \mathbb{R}^{P\times D}$.

Learnable \textit{Spatio-temporal Position Embeddings} and \textit{Modality Type Embeddings} are added to each modality to retain both positional and modality information.
Position embedding $V^{pos}\in \mathbb{R}^{(N+1)\times D}$ and text positional embedding is $T^{pos} \in \mathbb{R}^{(P+1)\times D}$.
The text and video embedding are further summed with their corresponding modal-type embedding vectors $t^{type},v^{type} \in \mathbb{R}^D$.
Formally,

\begin{equation}
\begin{aligned}
\hat{t}=[t_{class};t_1 T;\dots;t_P T] + T^{pos} + t^{type} \\
\hat{v}=[v_{class};v_1V;\dots;v_N V] + V^{pos} + v^{type} \\
\end{aligned}
\end{equation}

These text tokens are connected in series with vision tokens of each frame and the joint input is recorded as $z^0 = [\hat{t}; \hat{v}]$.
Then $z^0$ is fed into $L$ stacked blocks and each block consists of a temporal Token Rolling layer, a multi-head self-attention layer and a multilayer perceptron (MLP).
We initialize the weights of both self-attention and MLP layers from pre-trained ViT \cite{vit} or DeiT \cite{deit}.
The visual features of each sampled frame are independently encoded using a visual backbone model to extract the relationship between the frame and its associated textual representation.
Formally, 

\begin{equation}
\begin{aligned}
z^{d-1} = TTR(z^{d-1}), d=1 ... L \\ z^d = MLP(MSA(z^{d-1})), d=1 ... L
\end{aligned}
\end{equation}
where $MSA$ means multiheaded
self-attention, $MLP$ is multilayer perceptron and $TTR$ is short for Temporal Token Rolling Module.
Independent predictions from all the sampled frames are fused together to derive a consensus at the video level. 
Formally, $p=\frac{1}{S}\sum_{i=1}^{S}z_i^{L}$.
For pre-training, objectives are calculated based on this consensus to learn model parameters.

\begin{figure}[t]
    \centering\small
    \begin{minipage}[t!]{0.47\textwidth}
        \centering\small
        \includegraphics[width=.85\columnwidth]{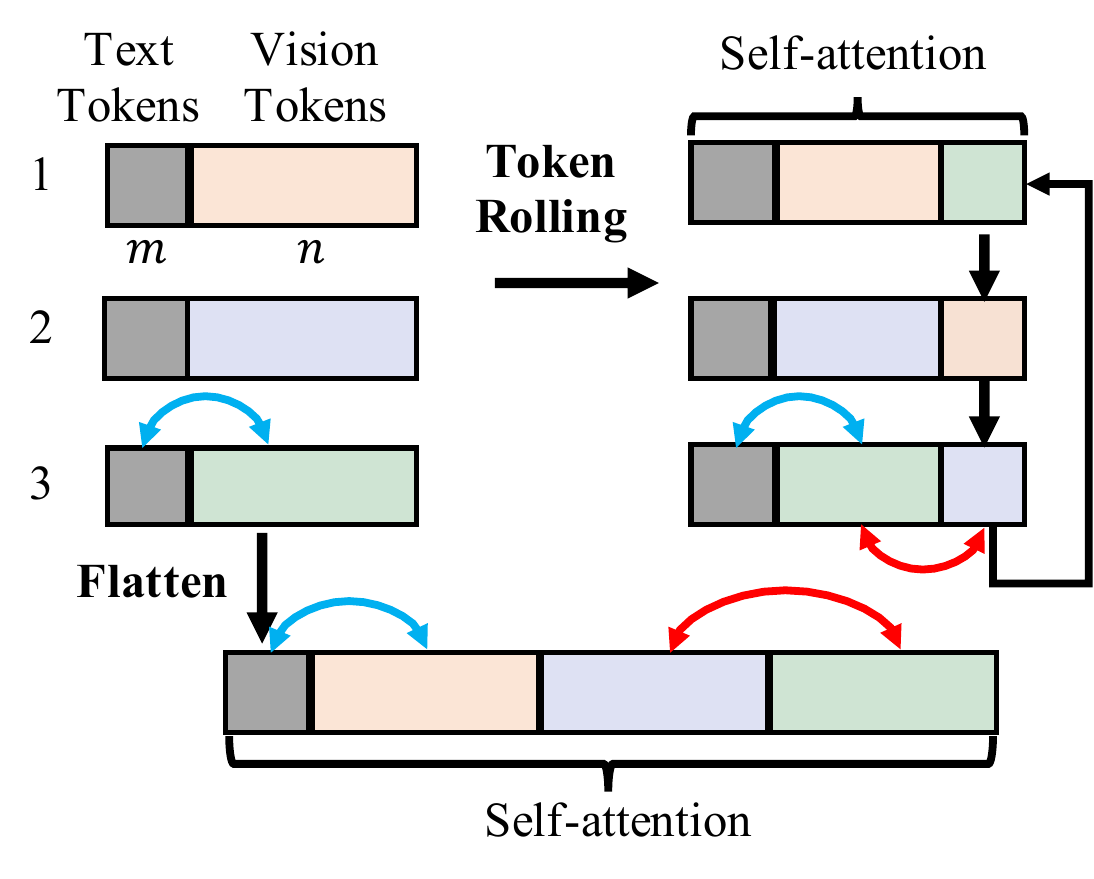}
        \vspace{-1em}
        \caption{%
        \textbf{The token rolling vs. flatten.}
        By simply rolling tokens, the computation complex for Self-attention is around one third of Flatten.
        Not only learns correspondence \textcolor{cyan}{cross-modality} but also \textcolor{red}{inter-frame}.
        %By
        %And show better results.
        %\xl{explain what purple and green means will be clearer}
        }
        \label{fig:3_token_roll_layer}
    \end{minipage}
    \hskip1em
    \begin{minipage}[t!]{0.47\textwidth}
        \centering\small
        \includegraphics[width=.85\columnwidth]{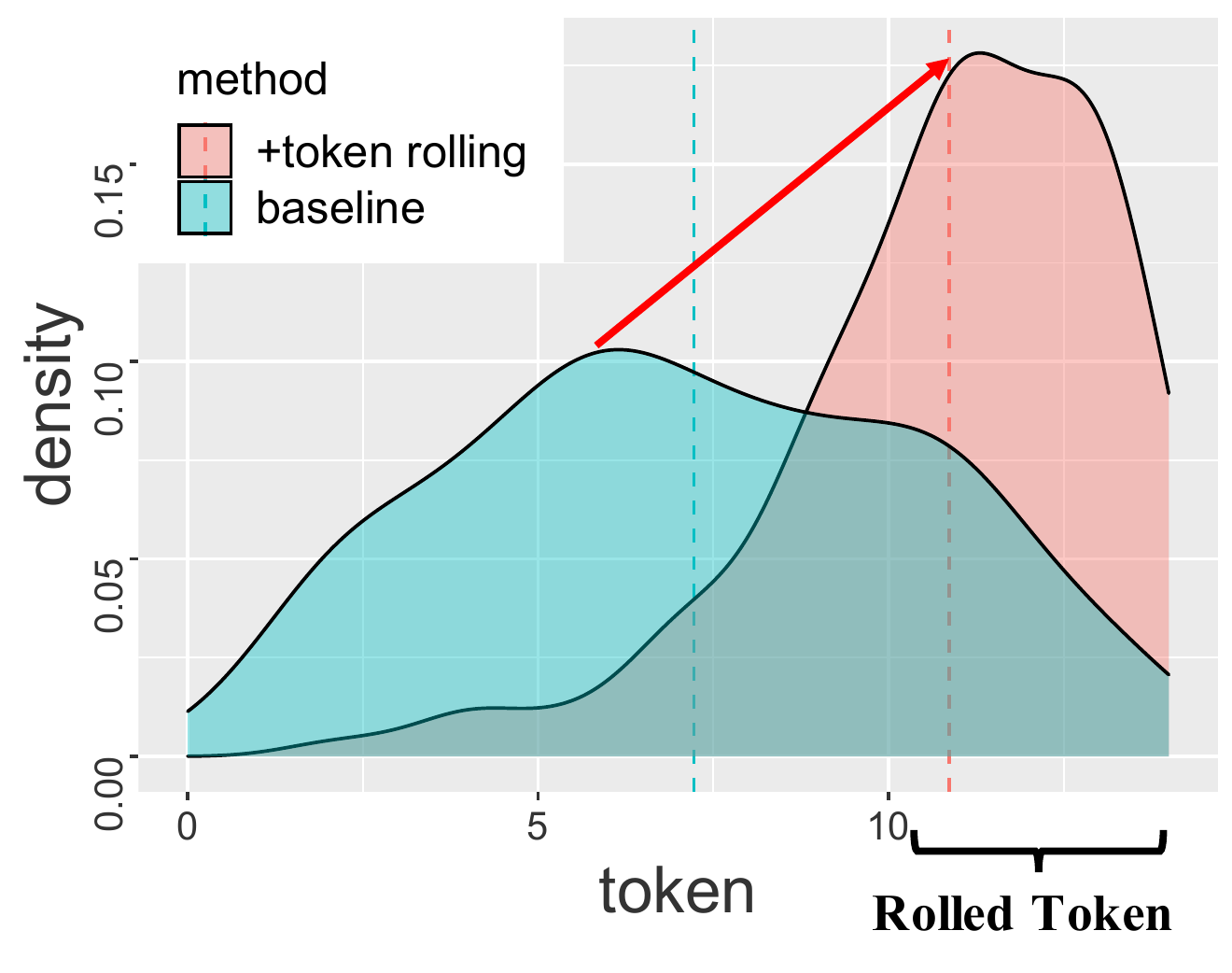}
        %\vspace{-1em}
        \vspace{-1em}
        \caption{%
        \textbf{The text to video attention weight distribution over tokens}.
        With the Temporal Token Rolling layer, the text token pay more attention to rolled tokens, in contrast to previous centric attention.
        }
        \label{fig:3_token_roll_weight_distribution}
    \end{minipage}
    \vspace{-1em}
\end{figure}

\subsection{Temporal Token Rolling}
\label{sec:ttr}

\textbf{Motivation.}
In VLP, the common usage for temporal modeling is to add additional time attention layers \cite{frozen} in vision encoder or use the feature from deep off-the-shelf video encoder \cite{actbert,violet}.
However, these techniques are particularly designed for video and thus can not be applied to process text signal, as well as bringing a large amount of parameters.
For example, by simply adding a temporal attention layer to each block of the Transformer, the model becomes a normal Timesformer \cite{timesformer} with parameters increased from 86M to 121.7M (an increase of 42\%).
Thus, these techniques cannot be used in our unified framework and we turn to find new ways to learn temporal information with modest parameters.

\noindent \textbf{Approach.} A straightforward approach, denoted as ``Flatten'', is to concatenate video and text tokens together and \textit{flatten} into one tensor which will be fed into the self-attention blocks.
Given text token with length $m$ and video token with length $S \times n$, we show the \textit{flatten} version in Fig.\ref{fig:3_token_roll_layer}.
However, as the self-attention layer has quadratic complexity, the computation cost will be $\mathcal{O}{((m+Sn)^2)}$, about $S^2$ times more than 1-frame \ModelName \footnote{The length of text tokens $m$ much smaller than video tokens $n$ in general.}.
To overcome such limitation, \textit{we try to exchange information for different time segments in the token level. }
The proposed Token Rolling module is described in Fig. \ref{fig:3_ppl} (b). 
The tokens at different time stamps are denoted as different colors in each row.
Along the temporal dimension, we roll parts of the token by 1, leaving the rest unchanged.
Then the self-attention is computed in each $m + n$ tokens and treat each token in the same way.
In this way, we reduce the computation complex to $ \mathcal {O}{(S(m+n)^2)}$, around $\frac{1}{S}$ of the Flatten version.

Taking advantage of Token Rolling, longer dependencies between texts and videos are gradually modeled in deeper layers, which helps to learn better video-text alignment.
We try to visualize the cross-modality attention weight density among text and video tokens in Fig. \ref{fig:3_token_roll_weight_distribution}.
For each text token, we compute the similarity by dot product to reveal its corresponding high-weight video tokens (more details are given in appendix).
The baseline is \ModelName without rolling layers.
We observe a severe inductive bias in the baseline, i.e., text tokens pay more attention to the centric tokens.
By introducing the Temporal Token Rolling, these rolled tokens contribute more to the cross-modality interaction.

\subsection{Training Objectives}
We train \ModelName with two objectives commonly used to train VLP models: video-text matching (VTM), and masked language modeling (MLM).
In addition, in order to overcome the disadvantage of low retrieval efficiency of one-stream, we introduce the video-text contrastive loss (VTC).

\textbf{Video-text Matching.}
Given a paired video-text input, we randomly replace the paired video with a different video with the probability of 0.5 and ask the model to distinguish them. 
For the \textit{cls token} of the last block, a single linear layer VTM head projects tokens to logits over binary class.
We compute negative log-likelihood loss as our VTM loss.

\textbf{Masked Language Modeling.}
MLM \cite{bert} aims to predict the ground truth labels of masked text tokens from the other text and video tokens.
Following the common practices \cite{bert,vilt}, we randomly mask text tokens with the probability of 0.15 and model it as a classification task.

\textbf{Video-text Contrastive.}
Inspired by the recent success of contrastive learning in visual-language pre-training \cite{frozen}, we also introduce this loss to our unified frameworks when fine-tuning for downstream video-text retrieval task.
Specifically, for video-text pairs, we input video and text independently to the shared encoder to obtain high-level features.
We then feed these features into a modality-specific projection head to project them into the shared embedding space. 
Following common practice \cite{frozen,oatrans}, we use a symmetric (both text-to-video and video-to-text) contrastive loss  based on these features.
When doing retrieval tasks, we only need to extract unimodal features once.

\section{Experiments}

To explore model scalability, we use large-scale Webvid-2.5M \cite{frozen}, HowTo100M \cite{howto100m} and YT-Temporal 180M \cite{merlot} for Pre-training.
We evaluate \ModelName on four popular downstream video-language tasks: text-to-video retrieval, video question answering, multiple-choice and visual commonsense reasoning across 9 different datasets.
We also provide extensive ablation studies to analyze the key factors that contribute to \ModelName's success, with insights and qualitative results.
More tasks and datasets are reported in the appendix.

%\vspace{-2em}
\subsection{Setup}

\subsubsection{Model Variants.}
When considering the generality of \ModelName, we consider using three configurations  based on ViT\cite{vit} and DeiT\cite{deit}, as summarized in Tab. \ref{tab:models}.
To simplify, we use brief notation to indicate the model size: for instance, \ModelNameBase/16 means the ``Base'' variant with $16 \times 16$ input patch size.
Following ViLT \cite{vilt}, we use the \textit{bert-base-uncased} tokenizer \cite{bert} to tokenize text inputs.
We random sample $3$ frames and resize each frame to $224 \times 224$. 
Then the patch projection of \ModelName yields $14 \times 14 = 196$ patches for each frame. 

\begin{table}[t]
\smallskip
\centering
\scalebox{.8}
{
\begin{tabular}{l|ccccccccc}
\toprule
Model       & Base & Initialization     & Embedding  & \#Heads & \#Layers & \#Params & Training & Throughput  \\

            & Model    &           & Dimension  &         &           &          & Resolution & (videos/sec)\\
\midrule
\ModelNameTiny  &  DeiT \cite{deit} &ImNet-21K &  \pzo192  & \pzo3 & 12 & \pzo12M & 224 & \pzo745 \\
\ModelNameSmall &  DeiT \cite{deit} &ImNet-21K& \pzo384   & \pzo6 & 12 & \pzo33M & 224 & \pzo285\\
% \ModelNameBaseDeiT  &  DeiT \cite{deit} &ImageNet-21K& \pzo768   & 12 & 12 & \pzo110M & 224 & \pzo89\\
\ModelNameBase  &  ViT \cite{vit}  & ImNet& \pzo768  & 12 & 12 & \pzo110M & 224 & \pzo89 \\
% \ModelNameLarge  &  ViT \cite{vit} &ImageNet& \pzo1024   & 12 & 12 & \pzo312M & 224 & \pzo30\\
\bottomrule
\end{tabular}}%}
\caption{Variants of our \ModelName architecture.
The throughput is measured for videos at a resolution of 224$\times$224. 
We use \ModelNameBase as default without specific explanation.
The ImNet is short for ImageNet.
\label{tab:models}}
\vspace{-2em}
\end{table}

\subsubsection{Pre-training \& Fine-tuning.}
Considering YT-Temporal 180M \cite{merlot} partially overlaps with HowTo100M \cite{howto100m}, we pretrain on WebVid2.5M+Howto100M as default.
If the model is trained on all three datasets, we named it as \ModelName*.
Due to the storage limitation, we use the first half of the YT-Temporal 180M.
We train all models using AdamW \cite{adaw} optimizer with a base learning rate of $10^{-4}$ and weight decay of $10^{-2}$. 
The learning rate was warmed up for 10\% of the total training steps and was decayed linearly to zero for the rest of the training. 

For pre-training, we train \ModelNameSmall and \ModelNameBase for 200K steps on NVIDIA A100 GPUs with a batch size of 16 per GPU \footnote{32 GPUs take 7 days total, 128 GPUs take less than 2 days total.}.
For \ModelNameTiny, we pre-train for 100K steps with a batch size of 32 per GPU, as we found it converges very fast.
We adopt mixed precision technique \cite{mixedtraining} to speed up the training process.
As the domain gap between pre-train dataset and downstream visual commonsense reasoning dataset is large, we use batch size 512 and train for 100 epochs for this task.
For the other downstream tasks, we train for 20 epochs with a batch size of 256.
Note that downstream performance may be further improved if we customize the hyperparameters to each task.

\subsection{Downstream Tasks}

\subsubsection{Video-question Answering.} 
\textit{Datasets:} In this work, we explore TGIF-QA \cite{tgifqa}, MSRVTT-QA \cite{msrvttqamsvdqa} and MSVD-QA \cite{msrvttqamsvdqa}.
We experiment with 3 TGIF-QA tasks: Repeating Action and State Transition for multiple-choice QA, and Frame QA for open-ended QA.
Both MSRVTT-QA and MSVD-QA are open-ended VQA.
\textit{Evaluation:} 
VQA requires answering questions according to the context of the video.
For open-ended VQA, the answers are originally in free-form natural language, but it is a common practice to convert the task to a classification task by representing the answer with a class label.
Following this practice, we add a two-layer MLP with hidden size 768 on the \textit{cls} token.
For multiple choice VQA (both the questions and candidates are sentences), we concatenate the question and candidates together, with [\textit{SEP}] to distinguish them.
We select the candidate with maximum output logit of VTM head as prediction.

\subsubsection{Text-video Retrieval.}
\textit{Datasets:} MSRVTT \cite{msrvtt}, DiDeMo \cite{didemo}, ActivityNet Captions \cite{activitynet_caption} are utilized for this task.
\textit{Evaluation.}
We initialize the similarity score head from the pre-trained ITM head, particularly the part that computes the true-pair logits.
We train this task with both VTC and VTM since we find these two objectives boost each other.
During inference, we simply feed each modality input independently and match pairs according to the cosine similarity of the output feature.
In this way, we take advantage of high efficiency of dual-stream frameworks in retrieval.

\subsubsection{Multiple-choice.}
\textit{Datasets:}
In this task, we adopt MSRVTT multiple-choice test set \cite{msrvttqamsvdqa} and LSMDC multiple-choice test set \cite{lsmdcmc}.
\textit{Evaluation:}
Given a video query and 5 candidate captions, the task is to find the one that fits the query out of 5 possible candidates. 
The correct answer is the ground-truth (GT) caption, and four other negatives
are chosen from other captions that have different activity-phrase labels from the correct answer.
We initialize the VTM head from the pretrained model on-top of the CLS token.
During the train, we simply concat each candidate with the given video together as input and only the correct answer is positive pair while the others negative pairs. 
We tune the model with cross-entropy loss to maximize the scores on positive pairs.

\subsubsection{Visual Commonsense Reasoning.} 
VCR \cite{vcr} is a task and dataset where models must answer commonsense visual questions about images. 
This task test our model's ability to transfer its video-level understanding to single image.
To solve this challenge task, VCR provides additional information to models (in the form of bounding boxes around entities), and explicit groundings between those entities and references in questions. 
Following previous efforts \cite{merlot}, we incorporate the location and identity information by drawing mask around the referenced entity.
%We finetune the model with different source.

\newcolumntype{g}{>{\columncolor{Gray}}c}
\begin{table*}[!t]
\small
\smallskip
\centering

\subfloat[Three sub-tasks on TGIF-QA test set (the first row are methods w/o. pre-training). ``\textit{T}'' refers to text encoder, ``\textit{V}'' is video encoder and ``\textit{CE}'' is cross-modality encoder.
384 means the resolution is $384\times 384$ for each frame while the default is $224\times 224$.
\label{tab:main_tgif_qa}]{
\scalebox{\GlobalTableRescale}
{
\begin{tabular}{l|cccc|ccg}
\toprule
Method&Nets&Params&PT Samples& Frames & Action &Transition & FrameQA \\
\midrule
%PSAC~\cite{psac}&&-&35& 70.4 & 76.9 & 55.7 \\
Heterogeneous~\cite{heterogeneous}&$\textit{T}$+$\textit{V}$+$\textit{LSTM}$&-&-& 35& 73.9 & 77.8 & 53.8 \\
HCRN~\cite{hcrn}&$\textit{T}$+$\textit{V}$+$\textit{LSTM}$&-&-&16& 75.0 & 81.4 & 55.9 \\
QueST~\cite{quest}&$\textit{T}$+$\textit{V}$+$\textit{LSTM}$&- &-&16& 75.9 & 81.0 & 59.7 \\
\midrule
ClipBERT \cite{clipbert} &$\textit{T}$+$\textit{V}$+$\textit{CE}$&137M&5.6M& $1 \times 1$ & 82.9 &  87.5 & 59.4 \\
VIOLET \cite{violet}&$\textit{T}$+$\textit{V}$+$\textit{CE}$&198M&5.5M&16&87.1&93.6&-\\
% \hline
\midrule
\ModelNameTiny&$\textit{CE}$&12M&3.72M& 3 & 80.6&83.5 & 53.9 \\
\ModelNameSmall&$\textit{CE}$&33M&3.72M& 3 & 91.2&92.7 & 64.0 \\
\ModelNameBase&$\textit{CE}$&110M&3.72M&  1 & \textbf{92.9} \greenp{5.8$\uparrow$}& 94.2 \greenp{0.6$\uparrow$}&  62.5 \greenp{3.1$\uparrow$}\\
\ModelNameBase &$\textit{CE}$&110M&3.72M& 3 & 92.7 \greenp{5.6$\uparrow$}& \textbf{94.3} \greenp{0.7$\uparrow$}&  \textbf{64.2} \greenp{4.8$\uparrow$}\\
\midrule
\demph{\ModelNameBase~[384]}&\demph{$\textit{CE}$}&\demph{110M}&\demph{3.72M}& \demph{3} & \demph{94.7}& \demph{95.1}&\demph{65.4}\\
\demph{\ModelNameBase*}&\demph{$\textit{CE}$}&\demph{110M}&\demph{9.72M}& \demph{3} &\demph{95.5} & \demph{94.7}& \demph{66.3} \\
\bottomrule
%T$^2$ST (384) & 3& -&  -\\
% \ModelName~1\x1 & \bf 82.8 & \bf 87.8 & \bf 60.3 \\
\end{tabular}
}
}
%\hspace{2mm}

\subfloat[MRSVTT-QA test set.
\label{tab:main_msrvtt_qa}]{
\scalebox{\GlobalTableRescale}
{
\begin{tabular}{l|c|g}
\toprule
Method & Frames & Accuracy \\
\midrule
AMU~\cite{amu} &16& 32.5 \\
Heterogeneous~\cite{heterogeneous}& 35& 33.0 \\
HCRN~\cite{hcrn} &16& 35.6 \\
ClipBERT \cite{clipbert}&$4 \times 2$&37.4\\
VIOLET \cite{violet}&16&43.1\\
% \hline
\midrule
\ModelNameSmall &3& 39.5  \\
\ModelNameBase   &3& 42.9 \redp{0.2$\downarrow$} \\
\ModelNameBase  &$3 \times 3$& \textbf{44.3} \greenp{1.2$\uparrow$} \\
\midrule
%\demph{\ModelName-B~[384]}&\demph{3} & \\
\demph{\ModelNameBase*} &\demph{3}&  \demph{46.8}  \\
%\multicolumn{3}{c}{} \\
\bottomrule
\end{tabular}
}
}
%\hspace{2mm}
\subfloat[MSVD-QA test set. \label{tab:main_msrvtt_multiple_choice_test}]{
\scalebox{\GlobalTableRescale}
{
\begin{tabular}{l|c|g}
\toprule
Method &  Frames&  Accuracy \\
% \shline
\midrule
QueST~\cite{quest} &10& 36.1 \\
HCRN~\cite{hcrn}& 16& 36.1\\
SSML~\cite{ssml} &16& 35.1 \\
CoMVT~\cite{comvt} &30& 42.6 \\
Just-Ask $\dag$ \cite{justask} &32&  46.3\\
%\hline
\midrule
\ModelNameSmall &3&  41.7 \\
\ModelNameBase &3& 46.5 \greenp{0.2$\uparrow$} \\
\ModelNameBase &$3 \times 3$& \textbf{47.9} \greenp{1.6$\uparrow$} \\
\midrule
%\demph{\ModelNameBase~[384]} &\demph{3}&  \\
\demph{\ModelNameBase*} &\demph{3}&\demph{48.3} \\
\bottomrule
\end{tabular}
}
}
\caption{
Comparison with state-of-the-art methods on VQA.
The columns with gray color are \textbf{open-ended VQA} and the others are \textbf{multiple-choice VQA}.
$\dag$ means use additional large-scale VQA dataset HowToVQA60M \cite{justask} for pretraining.
$^\star$ means pre-training with additional YT-Temporal 180M \cite{merlot}.
}
\label{tab:vqa_sota}
\vspace{-2em}
\end{table*}

\subsection{Comparing to State-of-the-art}

\subsubsection{Video-question Answering.}
In this experiment, we compare three variations of our \ModelName to state-of-the-art methods from the literature.
For multiple-choice VQA, we evaluate our \ModelName on two sub splits of TGIF-QA and report the result in Tab. \ref{tab:vqa_sota}.
We find \ModelName especially good at this type of VQA.
With ouly 1 frame input, our \ModelNameBase outperforms previous VIOLET \cite{violet} about 5.8\% on the Action subset.
Interestingly, we find more frames not benefit to Action and Transition but FrameQA.
We also report the result of \ModelName on the three open-ended datasets. 
Surprisingly, even though Just-Ask \cite{justask} is specifically designed for VQA and pretrined on large scale HowToVQA69M, our method still achieves a similar even better result than Just-Ask on MSVD-QA.

\begin{table}[t]
% \raggedright
\small
\centering
\begin{subtable}{\linewidth}\centering
{
\scalebox{\GlobalTableRescale}
{
\begin{tabular}{l|cccc|ccc|ccc}
\toprule
 Method & Nets & PT Data&  Params &  Frames & \multicolumn{3}{c|}{  9K Train} & \multicolumn{3}{c}{ 7K Train}\\ 
&&&&& R@1 & R@5 & R@10 & R@1 & R@5 & R@10 \\
%\shline
\midrule
% \demph{HERO~\cite{li2020hero} ASR, PT} && & \demph{20.5} & \demph{47.6} & \demph{60.9} & \demph{-} \\
ActBERT~\cite{actbert}&$\textit{T}$+$\textit{O}$+$\textit{V}$+$\textit{CE}$&HowTo&275M&32&-&-&-&16.3 & 42.8 & 56.9  \\
ClipBERT~\cite{clipbert} &$\textit{T}$+$\textit{V}$+$\textit{CE}$&COCO+VG&137M&$8 \times 2$&-&-&-& 22.0 & 46.8 &  59.9 \\
TACo~\cite{taco} & $\textit{T}$+$\textit{V}$+$\textit{CE}$&HowTo&212M& 48& 28.4&57.8&71.2&24.8&52.1&64.0 \\
% VideoCLIP &&& 30.9 &55.4 &66.8 \\
VIOLET \cite{violet} &$\textit{T}$+$\textit{V}$+$\textit{CE}$&CC+WebVid&198M&16& 34.5& 63.0&73.4&-&-&-\\
\demph{Frozen~\cite{frozen}}&\demph{\textit{T}+\textit{V}}&\demph{CC+WebVid}&\demph{232M}&\demph{8}& \demph{31.0} & \demph{59.5} & \demph{70.5} &\demph{-}&\demph{-}&\demph{-}\\
% \demph{RegionLearner~\cite{regionlearner}}&\demph{\textit{T}+\textit{V}}&\demph{CC+WebVid}&\demph{242M}&\demph{8}& \demph{36.3} & \demph{63.9} & \demph{ 72.5} &\demph{-}&\demph{-}&\demph{-}\\
\demph{OA-Trans~\cite{oatrans}}&\demph{\textit{T}+\textit{O}+\textit{V}} &\demph{CC+WebVid}&\demph{232M}&\demph{8}& \demph{35.8} & \demph{63.4} &\demph{76.5} &\demph{32.1} & \demph{61.0} &\demph{72.9}  \\
%\hline
\midrule
%\ModelNameSmall~&$\textit{CE}$&HowTo+WebVid&33M&$3$& \\
\ModelNameBase~&$\textit{CE}$&HowTo&110M&3&29.5&63.3&71.9&26.5&59.4&69.8\\
\ModelNameBase~&$\textit{CE}$&HowTo+WebVid&110M&3& 37.1 & 66.7& 75.9 & 33.8 & 64.2& 74.3\\
\ModelNameBase~&$\textit{CE}$&HowTo+WebVid&110M&$3 \times 3$&\bf  37.9 &\bf  68.1& \bf 77.1& \bf 34.4 & \bf 65.4& \bf 75.8 \\
%\midrule
%\ModelNameBase~[384]&$\textit{CE}$&HowTo+WebVid&110M&$3$& \\
% \ModelNameLarge~&$\textit{CE}$&330M&$3$& \\
\bottomrule
%\multicolumn{5}{c}{} \\
\end{tabular}
}
}
\caption{The retrieval performance on MSR-VTT 9K and 7K training split. For Nets, ``\textit{O}'' is object extractor.
HowTo is short for HowTo100M \cite{howto100m}.
Notice that COCO \cite{coco}, CC (short for Conceptual Captions \cite{cc3m}) and VG (short for Visual Genome \cite{visualgenome}) are all image-text datasets, which are not suitable for temporal modeling during pre-training.}
\label{tab:baseline_a}
\end{subtable}%

\begin{subtable}{.49\linewidth}
{
\scalebox{\GlobalTableRescale}
{
\begin{tabular}{l|c|rrrr}
\toprule
Method & Frames&  R@1 &  R@5 &  R@10 &MdR \\
\midrule
Dense~\cite{activitynet_caption}&32&14.0 & 32.0&-&34.0 \\
FSE~\cite{fse}&16&	18.2& 44.8	& - & 7.0 \\
HSE~\cite{fse}&8& 20.5 & 49.3  & - & - \\
ClipBERT\cite{clipbert}&$4 \times 2$&20.9& 48.6& 62.8& 6.0 \\
%\hline
\midrule
% \ModelNameSmall~ &3\\
\ModelNameBase~ &3& 21.5 & 50.3 & 65.5 & 6.0\\
\ModelNameBase~ &$3 \times 3$&\textbf{22.4} & \bf 53.7 &\bf 67.7 & \bf 5.0\\
\bottomrule
\end{tabular}
}
}
\caption{ActivityNet Caption val1 set.}
\label{tab:baseline_b}
\end{subtable}
\begin{subtable}{.49\linewidth}\centering
{
\scalebox{\GlobalTableRescale}
{
\begin{tabular}{l|c|rrrr}
\toprule
Method & Frames& R1 &  R5 &  R10 & MdR \\
%\shline
\midrule
FSE~\cite{fse}	&16& 13.9	& 36.0 & -&11.0  \\
CE~\cite{ce} 	&16& 16.1	&  41.1 & - & 8.3 \\
ClipBERT \cite{clipbert} &$8 \times 2$ & 20.4 & 48.0 & 60.8&6.0  \\
% \rowcolor{LightCyan}
\demph{Frozen \cite{frozen}} &\demph{8}& \demph{31.0} & \demph{59.8}& \demph{72.4}& \demph{3.0}\\
%\hline
\midrule
% \ModelNameSmall~&3& \\
\ModelNameBase~ &3& 31.2 & 60.5 & 72.1&3.0\\
\ModelNameBase~ &$3\times3$&  \bf 32.7 & \bf 61.4 & \bf 73.5&\bf 3.0\\
\bottomrule
\end{tabular}
}
}
\caption{DiDeMo test set.}
\label{tab:baseline_c}
\end{subtable}
\caption{Comparison with state-of-the-art methods on text-to-video retrieval. 
We gray out \demph{dual-stream} networks that only do retrieval tasks. Notice that OA-Trans \cite{oatrans} uses additional offline object features.}
\label{tab:retrieval_sota}
%\vspace{-3em}
\end{table}

\subsubsection{Retrieval Tasks.}
In this experiment, we fine-tune \ModelName on MSRVTT, ActivityNet Caption and DiDeMo datasets.
Tab. \ref{tab:retrieval_sota} summarizes results on text-to-video retrieval. 
In Tab. \ref{tab:retrieval_sota}(a), \ModelName achieves significant performance gain over existing methods on MSRVTT retrieval in both 9K and 7K train setting. 
Compre with these related works, we only use one Cross-modality Encoder and the parameter is half of the Frozen \cite{frozen}.
\ModelName even leads to 2.1\% relative improvement on R@1 when compare with OA-Trans \cite{oatrans}, which use additional offline object feature and only focus on retrieval.
When adopt on LSMDC and DiDeMo dataset, our method also show competitive result.

%\vspace{-1.1em}

\subsubsection{Multiple-choice.}
Tab. \ref{tab:mc} shows that \ModelName improves ClipBERT model by 3.2\% on accuracy, on MSRVTT multiple choice test task.
We also report the zero-shot performance for comparison, we find zero-shot accuracy already close to JSFusion \cite{jsfusion} in MSRVTT multiple choice with only three frames as input.

\begin{table}[t]
\raggedright
\centering
\begin{subtable}{.45\linewidth}
{
\scalebox{\GlobalTableRescale}
{
\begin{tabular}{l|c|c}
\toprule
Method &\ Frames& Accuracy  \\
%\shline
\midrule
JSFusion \cite{jsfusion} &40&83.4 \\
ActBERT \cite{actbert} &32&85.7 \\
% CLIPBERT \cite{clipbert}& 4$\times$1&  87.9 \\
ClipBERT \cite{clipbert}& $8 \times 2$&  88.2 \\
%\hline
\midrule
\ModelNameBase &3&  91.4  \\
\ModelNameBase &$3 \times 3$& \textbf{92.0} \greenp{3.8$\uparrow$}  \\
\midrule
\demph{\ModelNameBase~*} & \demph{3} & \demph{92.3} \\
\rowcolor{LightCyan} 
\ModelNameBase (zero-shot) &3& 80.3  \\
\bottomrule
\end{tabular}
}
}
\caption{MRSVTT multiple-choice test.}
\label{tab:mc_msrvtt}
\end{subtable}
\begin{subtable}{.45\linewidth}\centering
{
\scalebox{\GlobalTableRescale}
{
\begin{tabular}{l|c|c}
\toprule
Method&Frames & Accuracy\\
\midrule
%\shline
JSFusion \cite{jsfusion} &40 & 73.5 \\
MERLOT \cite{merlot} &8& 81.7	\\
VIOLET \cite{violet} &16& 82.9 \\
%\hline
\midrule
\ModelNameBase &3 & 83.1 \\
\ModelNameBase &$3 \times 3$& \textbf{83.5} \greenp{0.6$\uparrow$} \\
\midrule
\demph{\ModelNameBase~*} & \demph{3} & \demph{84.4} \\
\rowcolor{LightCyan}
\ModelNameBase (zero-shot)&3& 56.3  \\
\bottomrule
\end{tabular}
}
}
\caption{LSMDC multiple-choice test.}
\label{tab:mc_lsmdc}
\end{subtable}
\caption{Comparison with state-of-the-art methods on multiple-choice task.}
\label{tab:mc}
%\vspace{-1em}
\end{table}

\begin{figure}[t]
  \centering
    \includegraphics[width=.85\linewidth]{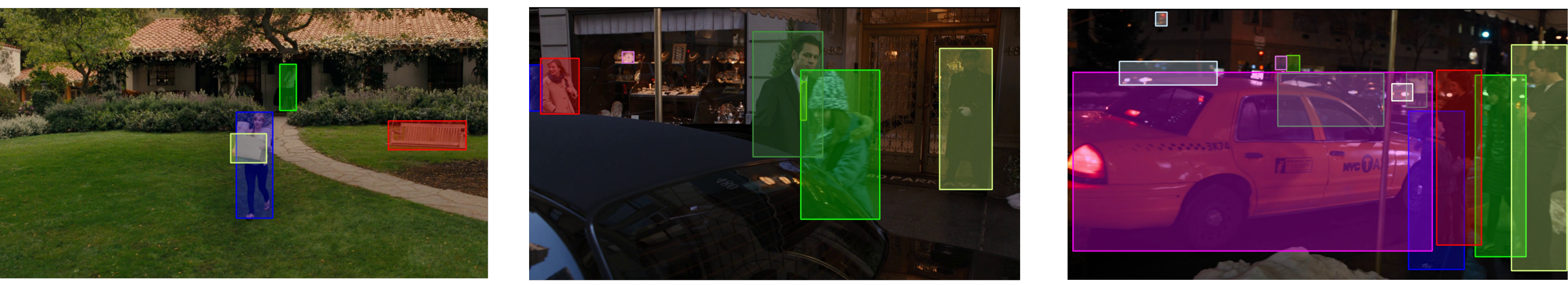}
   \caption{
    \textbf{Some examples from VCR \cite{vcr} dataset.}
    We add different colors around the identities to be consistent with the identities in the Q\&A text.
   }
   \label{fig:4_vcr_example}
   \vspace{-1em}
\end{figure}

%\newcolumntype{g}{>{\columncolor{Gray}}c}
\begin{table}
\footnotesize
\centering
\begin{minipage}[b]{0.49\hsize}
{
\scalebox{\GlobalTableRescale}
{
\begin{tabular}{l|cc|c}
\toprule
Method &PT Data&Mask&Accuracy \\
%\hline
\midrule
%\shline
MERLOT \cite{merlot}&CC3M+COCO & \textcolor{darkergreen}{\checkmark} & 58.9  \\
MERLOT \cite{merlot}& HowTo100M & \textcolor{darkergreen}{\checkmark} & 66.3  \\
%\hline
\midrule
\ModelNameBase & CC3M+COCO & \textcolor{darkergreen}{\checkmark}& \bf 60.5 \greenp{1.6$\uparrow$}\\
\ModelNameBase & HowTo100M && 65.2\\
\ModelNameBase & HowTo100M & \textcolor{darkergreen}{\checkmark}& \bf 68.4 \greenp{2.1$\uparrow$}\\
\bottomrule
\end{tabular}
}
\caption{
The visual commonsense reasoning result with different source of pre-training data.
}
\label{tab:vcr_sota}
}
\end{minipage}
\hskip1em
\begin{minipage}[b]{0.4\hsize}
{
\scalebox{\GlobalTableRescale}
{
\begin{tabular}{l|c|cc|c}
\toprule
Initialization&Steps & MLM & ITM &LSMDC\\
\midrule
- &200K & 58.7 & 89.8 & 51.4  \\
- &800K & \textbf{62.3} &  91.5 & 55.5  \\
\midrule
ImageNet&200K& 60.3 &  90.5 & 52.2   \\
ImageNet-21K&200K& 60.5 & 90.4 & 52.5  \\
ImageNet-21K&800K& 62.2 & \textbf{92.1} & \textbf{55.7} \\
\bottomrule
\end{tabular}
}
\caption{Comparison with different initialization for \ModelNameBase.}
\label{tab:abl_initalie}
}
\end{minipage}

\end{table}

\subsubsection{Visual Commonsense Reasoning.}
After pre-training, we use a visual reasoning task to test the generality ability of our model.
Our results on the VCR dataset, in comparison to other models at the same (``base'') scale, are given in Tab. \ref{tab:vcr_sota}.
Moreover, to utilize identity information, we also mask the different identity with different color (as shown in Fig. \ref{fig:4_vcr_example}).
We observe our model outperforms MERLOT clearly in the same setting with different sources of data.

\subsection{Analysis of Temporal Token Rolling}

\begin{table}
\footnotesize
\centering
\scalebox{\GlobalTableRescale}
{
\begin{tabular}{ll|cccccc|cccccc}
\toprule
VTM&VTC &\multicolumn{6}{c|}{MSR-VTT}&\multicolumn{6}{c}{DiDeMo} \\
% \hline
\midrule
&& \multicolumn{3}{c}{T2V}&\multicolumn{3}{c|}{V2T}& \multicolumn{3}{c}{T2V}&\multicolumn{3}{c}{V2T}\\
& &R@1&R@5&R@10&R@1&R@5&R@10&R@1&R@5&R@10&R@1&R@5&R@10 \\
%\shline
\midrule
%\shline
&\textcolor{darkergreen}{\checkmark}& 36.7& 62.6& 72.2& 36.4& 62.9& 71.9 & 29.6& 47.3& 56.67& 29.6& 48.0&57.2\\
\textcolor{darkergreen}{\checkmark}& & 33.3& \bf 67.0& \bf 76.9& 34.4& \bf 66.3& 76.6 & 30.2 & 57.5&68.7 &27.7&\bf 59.6& \bf 69.5 \\
 \textcolor{darkergreen}{\checkmark}&  \textcolor{darkergreen}{\checkmark} & \bf 37.1 & 66.7& 75.9&\bf 37.5 & 66.1& \bf 77.4 & \bf 31.2 & \bf 59.5 & \bf 72.1 & \bf 30.4 & 58.3 & 69.4 \\
\bottomrule
\end{tabular}
}
\caption{
The text-to-video retrieval (T2V) and video-to-text retrieval (V2T) results with different objectives on finetuning for retrieval.
}
\vspace{-1em}
\label{tab:abl_objective}
\end{table}

\begin{table}
\footnotesize
\centering
%\raggedright
\begin{subtable}{.5\linewidth}\centering
{
\scalebox{\GlobalTableRescale}
{
\begin{tabular}{l|cc|c|c}
\toprule
Method &PF&DF&LSMDC&MSR-VTT \\
%\shline
\midrule
% \multicolumn{3}{l}{\demph{Additional Parameter}} \\
% Time Attention & & \\
% Decouple Attention & & \\
%\shline
%\multicolumn{3}{l}{\demph{Parameter-free}} \\
%\multicolumn{3}{c}{\demph{1-frame pretrain}} \\
\demph{Single Frame} &\demph{1}&\demph{1}&\demph{32.78} & \demph{67.22} \\
\demph{Time Average} &\demph{1}&\demph{3}& \demph{35.16} & \demph{68.25}  \\
\midrule
%\hline
%\multicolumn{3}{c}{\demph{3-frame pretrain}}\\
Time Average $\ddag$ &3&3& 47.33 & 72.82 \\
Channel Shift \cite{tsm} &3&3& 48.24 & 73.16  \\
%\hline
Flatten &3&3 & 51.40 & 73.28 \\
Token Rolling &3&3& \bf 52.75& \bf 75.42\\
\bottomrule
\end{tabular}
}
}
\caption{Different variations of parameter-free temporal modeling operation. }
\end{subtable}
\begin{subtable}{.4\linewidth}\centering
{
\scalebox{\GlobalTableRescale}
{
\begin{tabular}{l|c|c}
\toprule
Ratio &LSMDC & MSR-VTT \\
%\shline
%\shline
\midrule
0 & 47.21  & 72.80  \\
0.1&48.52 \greenp{1.31$\uparrow$} &73.59 \greenp{0.79$\uparrow$}  \\
0.25& \textbf{52.90} \greenp{5.69$\uparrow$}& \textbf{ 76.44}  \greenp{3.64$\uparrow$}\\
0.5 & 50.47 \greenp{3.26$\uparrow$}& 74.89 \greenp{2.09$\uparrow$}\\
0.75 & 44.55 \redp{2.66$\downarrow$}&69.45 \redp{3.35$\downarrow$}\\
\bottomrule
\end{tabular}
}
}
\caption{The ratio of rolled tokens in Token Rolling layer.
Excessive ratio lead to unstable training.
}
\end{subtable}
\caption{
The ablation study on zero-shot mutiple-choice. PF is short for pre-training frames and DF is short for downstream frames.
}
\label{tab:abl_token_var}
\vspace{-2em}
\end{table}

\subsubsection{The Variations of Temporal Modeling.}
To better study Token Rolling, we also train our \ModelName in four different settings: Single Frame: Pre-training and inference with 1 frame.
Time Average: Pre-training with 1 frame but inference with 3 frames.
Time Average $\ddag$: Pre-training and inference with 3 frames.
Channel Shift: we replace each Token Rolling layer with channel shift operation.
Flatten: As presented in Sec. \ref{sec:ttr}.
We observe: \emph{i}. Pre-training with multiple frames are essential for SSL tasks. e.g, from 35.16 to 47.33 on LSMDC.
\emph{ii}. The Token Rolling boosted an amazing 5.42\% improvement over time average baseline. 
Compared with channel shift \cite{tsm}, the rolling on tokens also show superior performance. 
We guess VLP require learn alignment between patches and the channel operation will erase this boundary.
\emph{iii}. Even the computation complex of Flatten is three times as Token Rolling, the performance of Token Rolling is slightly better than Flatten in both setting.
As we discussed in Sec. \ref{sec:ttr}, the benefits might come from the Token Rolling is a natural extension of self-attention among patches.

\subsubsection{Ablation on the Rolling Token Ratio.}
In order to understand how many tokens are needed to roll during pre-training, we conduct an ablation study as indicated in Tab. \ref{tab:abl_token_var} (b).
We follow the \ModelName protocol in the pre-training setup except for a smaller 1024 batch size and 100K steps.
Compared to the temporal average baseline (ratio equals to 0), we observe an amazing 5.69\% improvement.
The benefits come from more effective temporal modeling.

\subsubsection{The Variations of Initialization.}
To answer the question if initialization is crucial for large-scale VLP. 
We initialize \ModelName with three versions: Scratch, ImageNet and ImageNet-21K.
We report the results in Tab. \ref{tab:abl_initalie} with different train iterations and make following observations:
\emph{i}. Train from scratch convergence slower than train from ImageNet pretrained model.
\emph{ii}. The combination of Webvid2.5M and Howto100M is large enough to train the model from scratch.
When training our \ModelName for 800K steps, we find the train from scratch is close to the ImageNet-21K initialized version in both Pre-training and downstream evaluation.

\subsection{Objectives of Retrieval}
To study the effect of objectives during fine-tuning, we experiment with three different combination of objectives.
As shown in Tab. \ref{tab:abl_objective}, the VTC loss have high R@1 and VTM is more effective in R@5 and R@10.
With the combination of three objectives, \ModelName can achieve best performance in all measurement.
Notice that these methods are trained in a dual-stream way.

\subsection{Visualization}
\begin{figure}[t]
  \centering
    \includegraphics[width=\linewidth]{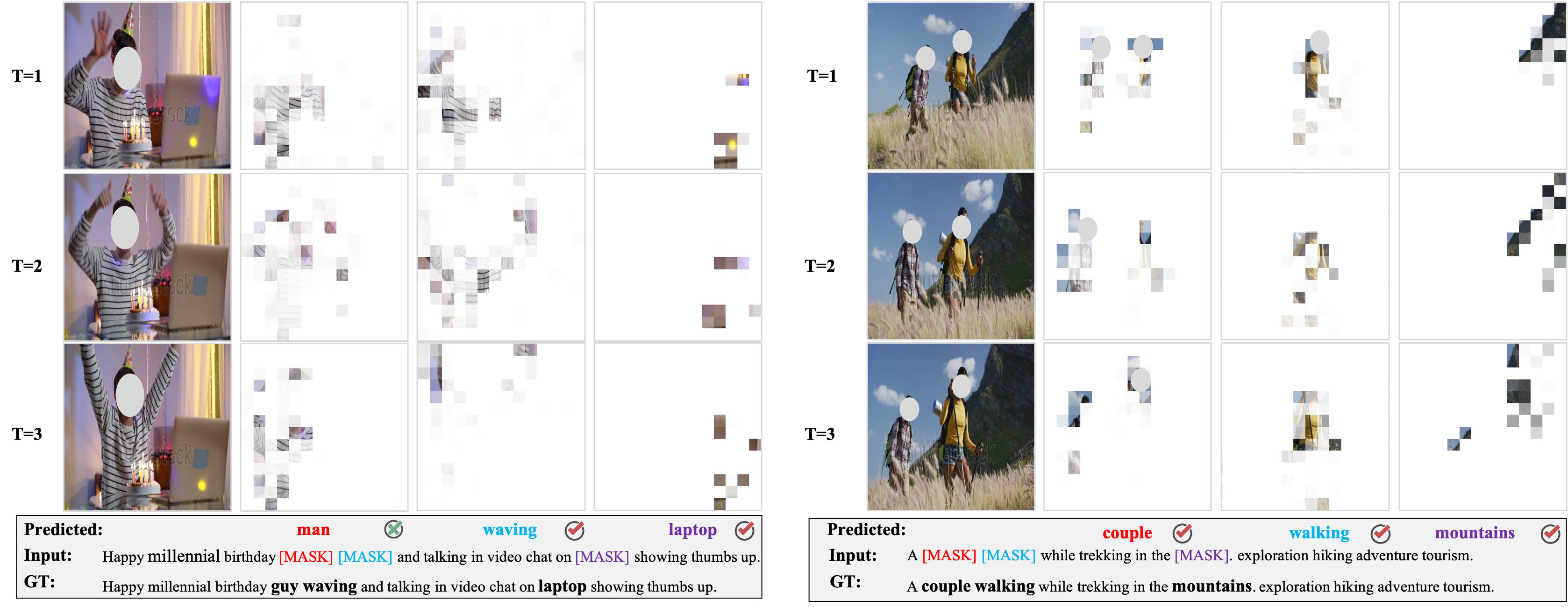}
   \caption{
   \textbf{\textit{Cloze} evaluation:} Given a video and its paired masked text, the model is asked to \textit{fill the masked words and show its corresponding high attention patch for this masked word.} 
   These samples are randomly sampled from the validation set of Webvid dataset \cite{frozen}.
   }
   \label{fig:4_attn_visualization}
%   \vspace{-2em}
\end{figure}

To better understand the pre-trained \ModelName, we analyze its internal representations.
Specifically, given paired ground truth text and raw video, we mask some keywords (both \textit{verb} and \textit{nouns}) and ask the model to predict these masked words and further find out which video patch has strong correlations with the masked words.
We use optimal transports \cite{ipot} to calculate the correlation between video and text.
We only show the attention weight that is larger than the given threshold and give some examples of cross-modal alignment in Fig. \ref{fig:4_attn_visualization}.
We find the model can predict correct \textit{nouns} and \textit{verbs} in most cases.
Sometimes, it predicts the wrong word but with a similar meaning to the correct word. e.g. ``guy'' and ``man''.
Benefiting from temporal modeling, we also find that the model attends to the motion regions for \textit{verbs} like ``waving'' and ``walking''.

\section{Conclusions}
In this paper, we present the first unified end-to-end Video-Language Pre-training architecture with raw video and text as input, \ModelName Transformer. 
By learning only one fusion network, \ModelName is able to complete with a large number of counterparts equipped with additional heavy off-the-shelf video visual embedding networks and holds promise for the future. 
We hope that the VLP community will focus more on lightweight end-to-end modal interactions within Transformer modules, rather than only on heavier single-modality embedders or larger fusion models.
While these initial results are encouraging, this new design of unified video-language interaction also brings new challenges, in particular fine-grained word region alignment. 
Furthermore, the temporal modeling is still not fully explored and we also encourage future work to use \ModelName for single-modality tasks.

\section*{Acknowledgement}
This project is supported by the National Research Foundation, Singapore under its NRFF award NRF-NRFF13-2021-0008.
We would like to thank David Junhao Zhang for his kindly help on Transformer training.

\begingroup
{
\bibliographystyle{splncs04}
\bibliography{reference}
}
\endgroup

%\newpage
\appendix
\section*{Appendix}

\appendix
%\tableofcontents

In this appendix, we first evaluate the model scalability and provide more ablation studies about \ModelName. 
Then we transfer \ModelName to more downstream tasks and datasets.
At last, we provide retrieval efficiency and more visualization analysis.

\section{Model Scalability}
%\vspace{-2em}
\begin{figure}[h]
    \centering
        \includegraphics[width=.95\columnwidth]{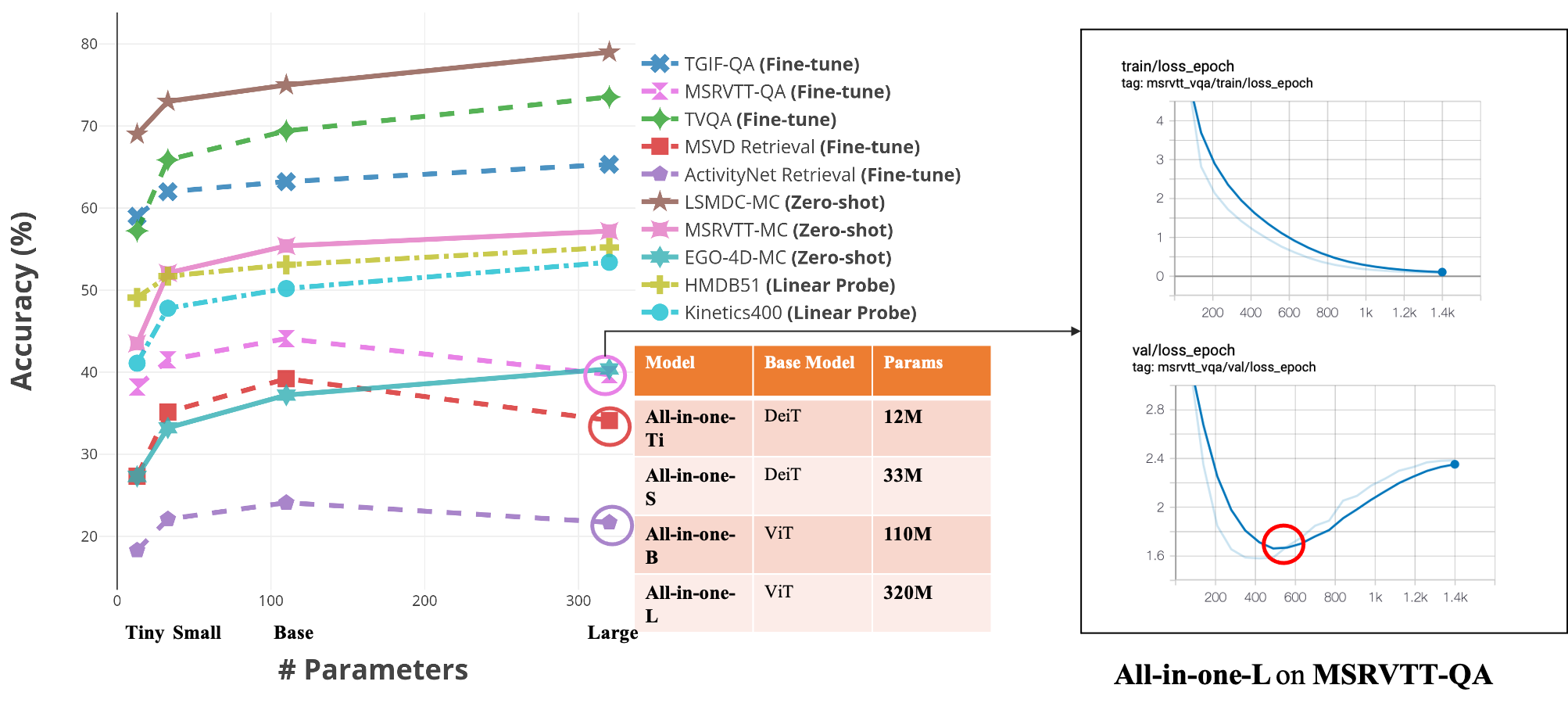}
        \caption{
        The parameters \& performance over ten downstream datasets with the varying of model size.
        }
        \label{fig:model_scalability}
\end{figure}

In this experiment, we evaluate the model from \ModelNameTiny to \ModelNameLarge using three assessment tasks: \textit{Fine-tune}, \textit{Zero-shot} and \textit{Linear Probe}. 
\textit{Zero-shot} means we directly test the pretrained model on downstream task without fine-tining, \textit{Linear Probe} means we frozen the overall model and only the last linear layer is learned on downstream tasks.
We varying the model size from 13M to 320M and do evaluation on 10 different datasets.
For fair comparison, the pre-training and fine-tuning settings are consistent for models of different scales.
The results are presented in Fig. \ref{fig:model_scalability} and we make the following observations:

\emph{i}. For \textit{Zero-shot} and \textit{Linear Probe} task, we observe large model leads to better result in general.
\emph{ii}.
However, we find sometimes \ModelNameLarge on the \textit{Fine-tune} task leads to worse result than \ModelNameBase in several benchmarks (circled).
We show the train curve of MSRVTT-QA in the right of Fig. \ref{fig:model_scalability}.
Since MSRVTT-QA only contains 10K video-text pairs, we find that the model severely overfits the dataset with few iterations.
For more large dataset like TVQA and TGIF-QA (ten times larger than MSRVTT-QA), \ModelNameLarge still lead to better results.
We conclude that simply pursuing larger models is not suitable for all cases, especially for fine-tuning on small scale, and \ModelNameBase is a better choice in most cases, as a trade-off between parameters and performance .

\section{Ablation Study}

In this work Masked Language Modeling (MLM) and Video-text Matching (VTM) are modeled as binary and multiple classes classification tasks, correspondingly. 
So we also report top-1 classification accuracy (\%) for these two pre-training objectives as reference in this section.

\subsection{The Variant of Temporal Modeling}

\begin{figure}[h]
    \centering
        \includegraphics[width=.95\columnwidth]{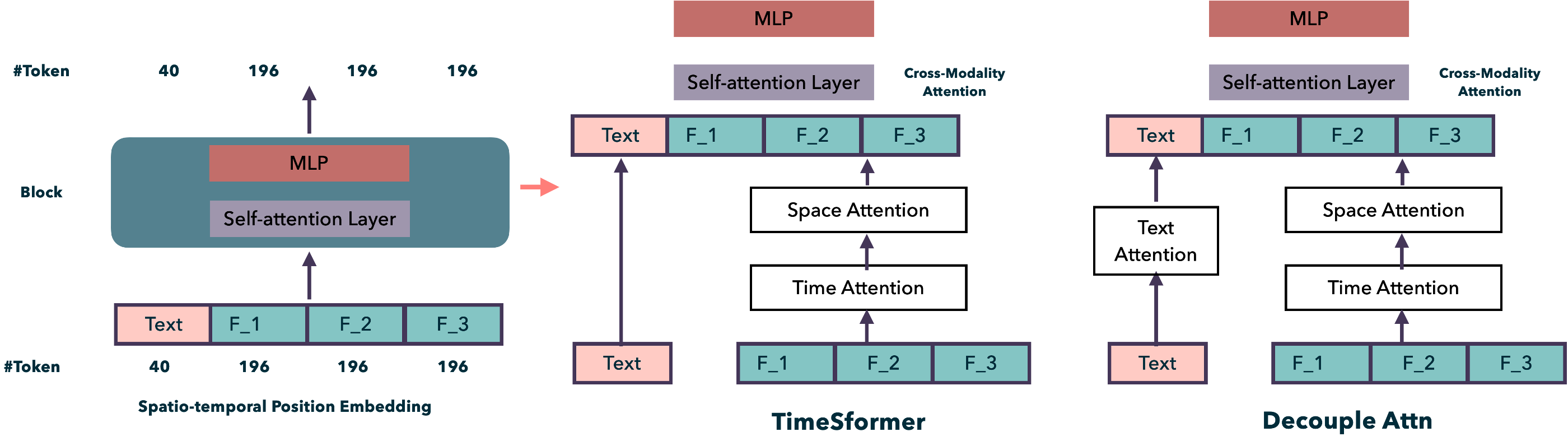}
        \caption{
        \textbf{TimeSformer} and \textbf{Decouple Attn} for temporal modeling.
        We mainly show the self-attention block of Transformer for simply.
        }
        \label{fig:time_attn}
\end{figure}

In addition to the parameter-free time token rolling operation proposed in this work, we also try different ways for temporal modeling:
\emph{i}. \textit{TimeSformer}: For each self-attention block in \ModelName, we add a additional divided space attention and time attention before multi-head self-attention layer. 
As shown in the middle of Fig. \ref{fig:time_attn}.
\emph{ii}. \textit{Decouple Attn}: For text modality and visual modality, we conduct self-attention independently first and then concatenate them again for cross-modality attention.
As shown in the right of Fig. \ref{fig:time_attn}.

Pre-training on Webvid2.5M + HowTo100M, we report both the pretrain and downstream evaluation performance in Tab. \ref{tab:supple_abl_temporal_attn}.
With more parameters in visual processing, we find that TimeSformer and Decouple Attn are particularly good at Video-text Matching, but not good at Masked Language Modeling.
However, we find that these methods are difficult to train, cost about 2-3 times more expensive than \ModelNameBase, and show worse results on downstream zero-shot tasks.

\newcolumntype{g}{>{\columncolor{Gray}}c}
\begin{table}[h]
\footnotesize
\centering
\scalebox{\GlobalTableRescale}{
\begin{tabular}{@{}llc|cc|cc@{}}
Method & Params & PT Time  & \multicolumn{2}{c|}{Pretrain} & \multicolumn{2}{c}{Downstream} \\

 &  &  & VTM& MLM  & LSMDC & MSR-VTT \\

\shline

TimeSformer&180.5M&75 Hours &91.5 &57.8& 53.2 &75.4   \\
Decouple Attn&192M&104 Hours &\textbf{92.3}&58.2& 52.1 &74.5  \\
\ModelNameBase&110M&32 Hours &90.5&\textbf{59.4}& \textbf{55.7} & \textbf{77.3}  \\
\end{tabular}
}
\caption{
\textbf{The variations of temporal modeling in the unified \ModelName}.
PT is short for pre-training and we report the zero-shot multiple-choice result.
% $\dag$ means add temporal positional embedding and temporal attention.
}
%\vspace{-2em}
\label{tab:supple_abl_temporal_attn}
\end{table}

\begin{figure}[h]
        \centering\small
        %\vskip0.8em
        \includegraphics[width=.35\columnwidth]{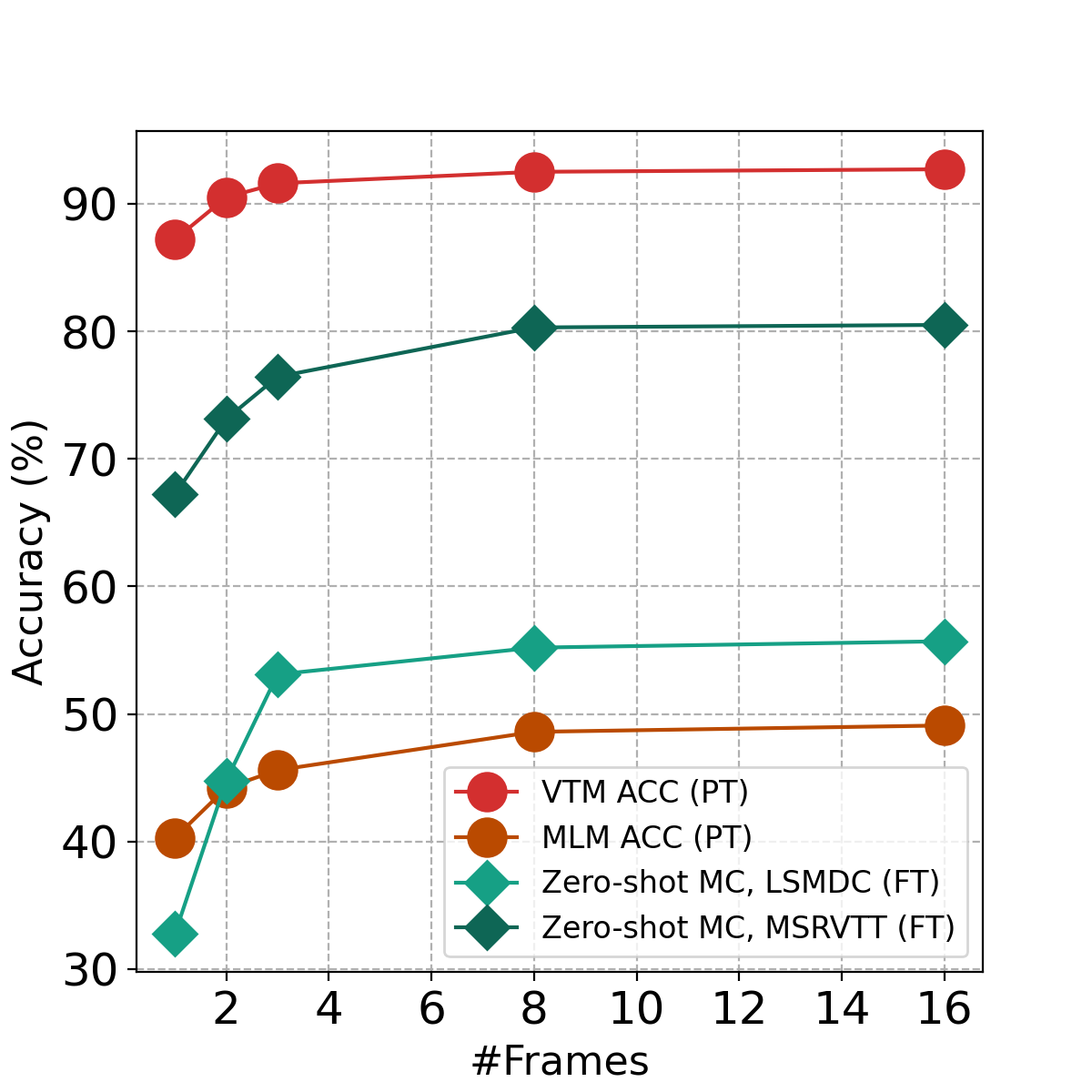}
        %\vskip0.8em
        \caption{%
        Both pretrain and finetune performance with \textbf{the varying of input frames}.
        %In this work, we set 3 as default for the balance of performance and efficiency.
        }
        \label{fig:frame_acc}
\end{figure}

\subsection{Do we need to sample more frames?} 
To further understand the relation between the number of frames and the quality of learned representation, we conduct pretrain and finetune experiments by varying the sampled frames for train.
Considering the memory consumption, we vary the frames from 1 to 16.
The difference between the default setting is that we pretrain on Webvid2.5M.
We report both the pretrain VTM and MLM accuracy and downstream zero-shot multiple-choice accuracy on both LSMDC and MSRVTT in Fig. \ref{fig:frame_acc}.
We find that more frames leads to better results in general and \ModelName is already close to the best results when frames equals to 3.
To balance the computation cost and performance, we use 3 frames as default.

\subsection{Position Embedding and Modality Type Embedding}
In this experiment, we explore the effect of Position Embedding and Modality Type Embedding.
The results are given in Tab. \ref{tab:abl_pos_embed}, we observe spatio-temporal position embedding help the VTM and modality type embedding helps the MLM pretext.
The combination of these two embedding leads to better results on both downstream zero-shot multiple choice result with limited parameters.

\newcolumntype{g}{>{\columncolor{Gray}}c}
\begin{table}[h]
\footnotesize
\centering
\scalebox{\GlobalTableRescale}{
\begin{tabular}{rr|rr|rr@{}}
SPE &MTE & \multicolumn{2}{c|}{Pre-training} & \multicolumn{2}{c}{Downstream} \\

& & VTM& MLM  & LSMDC & MSR-VTT \\

\shline
\redxmark &\redxmark &88.1 &51.0& 53.2 &74.4   \\
\textcolor{darkergreen}{\checkmark} &\redxmark &89.1&50.8& 54.5 &76.1 \\
\textcolor{darkergreen}{\checkmark}&\textcolor{darkergreen}{\checkmark} &90.4&52.3& 55.7 &78.2  \\
\end{tabular}
}
\caption{
The input analysis of \ModelNameBase.
SPE is short for Spatio-temporal Position Embedding and MTE is short for Modality Type Embedding.
}
\label{tab:abl_pos_embed}
\vspace{-2em}
\end{table}

\section{In-depth Analysis of Token Rolling}
\subsection{The Position of Temporal Token Rolling Layers}
In this experiment, we explore where to add our temporal rolling layer.
Specifically, \ModelNameBase contains 12 self-attention blocks and we add the Token Rolling Layers from the beginning, the 3rd block and the 6th block.
The results are report in the left of Tab. \ref{tab:abl_token_var}.
We observe more temporal token rolling layers leads to better representation.

\begin{table}[h]
\footnotesize
\centering
%\raggedright
\begin{subtable}{.4\linewidth}\centering
{
\scalebox{\GlobalTableRescale}
{
\begin{tabular}{l|c|c}
\toprule
Position&LSMDC&MSR-VTT \\
\midrule
1 & \textbf{56.15} & \textbf{76.16}  \\
3 & 55.34 & 75.16  \\
%\hline
6 & 54.40 & 73.31 \\
\bottomrule
\end{tabular}
}
}
\caption{The begin positions to add temporal token rolling layers. }
\end{subtable}
\begin{subtable}{.5\linewidth}\centering
{
\scalebox{\GlobalTableRescale}
{
\begin{tabular}{l|c|c}
\toprule
Method &LSMDC & MSR-VTT \\
%\shline
%\shline
\midrule
Random Selection & 54.88 & 75.03  \\
Varying Layers &50.83  &72.23  \\
Block& \textbf{55.81} & \textbf{ 76.37}  \\
\bottomrule
\end{tabular}
}
}
\caption{Different ways of token selection.
}
\end{subtable}
\caption{
The ablation study on zero-shot multiple-choice.
}
\label{tab:abl_token_variation}
%\vspace{-2em}
\end{table}

\subsection{The Sampling Strategy of Rolled Token}
In this experiment, we explore the sampling strategy of sampling token.
We try three versions:
\emph{i}.
Random Selection: Random select 25\% tokens.
\emph{ii}.
Varying with layers: For the first layer, we roll the first top 25\% and then the second top 25\%.
\emph{iii}.
Select a block that contains 25\% tokens.
As shown in the right of Tab. \ref{tab:abl_token_variation}, selecting the 25\% tokens leads to best result.
We guess this is due to the multiple perceptron is position sensitive and the random selection or varying layers will lost this information.
In this work, we adopt block selection as default.

\begin{table}[h]
\footnotesize
\centering
\scalebox{.9}{
\begin{tabular}{lcc|rrr|rrr}
\textbf{Method} &\textbf{Parameters} &\textbf{\#Frames}&\multicolumn{3}{c|}{\textbf{K400}}&\multicolumn{3}{c}{\textbf{HMDB51}}\\
&&&Top-1&Top-5&Top-10 &Top-1&Top-5&Top-10\\
\shline
Frozen \cite{frozen} & 232M & 8 & 50.5 & 80.7 & 90.2&54.3&88.0&\textbf{95.8}\\
Time Average&110M&3&44.3& 75.2& 87.3& 43.1 & 75.5 & 90.5   \\
\hline
\ModelNameBase&110M&3&50.8& 79.8 &90.7&52.9 & 84.1 & 93.4  \\
\ModelNameBase&110M&8&\textbf{53.4}& \textbf{83.2} &\textbf{92.6}&\textbf{55.7} & \textbf{88.2} & 95.2  \\
\end{tabular}
}
\caption{
The linear probe results on action recognition benchmarks with three frames over kinetics 400 and hmdb51 datasets.
}
%\vspace{-3em}
\vspace{-2em}
\label{tab:action_recognition}
\end{table}

\section{Transferability Evaluation}
\subsection{Action Recognition via Linear Probe}
To evaluate the transfer ability of our model on single-modality task. 
We transfer the learned representation to downstream linear probe result on K400 and HMDB51 dataset.
Specifically, we \textit{frozen} the overall unified model and only learns linear layers based on the \textit{cls} token of the last layer.
By pre-training model on these two datasets, we compare the base model with Time Average and previous best method Frozen \cite{frozen}.

The linear probe results are given in Tab. \ref{tab:action_recognition}.
We observe the number of frames have large impact on this task.
When adopt same 8 frames, our \ModelNameBase clearly outperforms Frozen \cite{frozen} especially on large-scale K400 dataset.

\begin{figure}[h]
    \centering
        \includegraphics[width=.8\columnwidth]{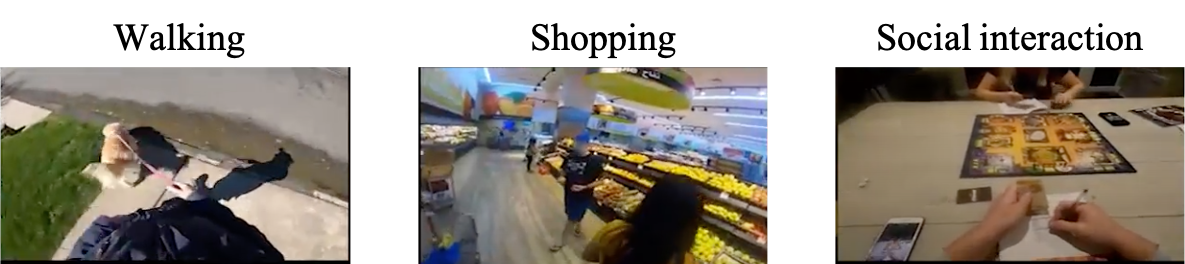}
        \caption{
        Some egocentric videos in Ego4d.
        }
        \label{fig:ego4d_examples}
\end{figure}

\subsection{Extension to Egocentric Video}

Ego-4d \cite{ego4d} is a egocentric dataset that has large domain gap with our third-view video from Youtube. 
We show some examples from Ego-4d in Fig. \ref{fig:ego4d_examples} and test multiple-choice (5 choices) task on this dataset.
We report both the zero-shot result and fine-tune result in Tab. \ref{tab:ego_4d}.
Compared to other multiple-choice benchmarks such as LSMDC and MSR-VTT, zero-shot accuracy is lower, but our \ModelName still outperforms Frozen \cite{frozen} clearly by half the parameters in this challenge benchmark.

\begin{table}[h]
\footnotesize
\centering
\scalebox{.9}{
\begin{tabular}{lcc|rr}
\textbf{Method} &\textbf{Parameters} &\textbf{\#Frames}&Zero-shot&Fine-tune \\
\shline
Frozen \cite{frozen} & 232M & 8 & 32.47 & 60.32\\
Time Average&110M&3& 27.34 & 59.44  \\
\hline
\ModelNameBase&110M&3& 36.52 & 65.89  \\
\end{tabular}
}
\caption{
The multiple-choice result on first-view ego-4d benchmark.
}
\label{tab:ego_4d}
\vspace{-2em}
\end{table}

\section{Complexity Analysis of Retrieval}
Due to the specifiy design for contrastive loss, Our \ModelName has a very fast inference running time even for retrieval on million-scale datasets. 
We use the popular similarity search/ranking library FAISS-GPU open-source library on a server with 8 A100 GPUs and 88 Kernel Intel(R) Xeon(R) Platinum 8255C CPU @ 2.50GHz.
Given a new query, Table \ref{tab:retrieval_time} below shows the time needed for visual encoding, textual encoding, and similarity ranking (1st row for thousand-scale and 2nd row for million-scale). 
Given a new query, the total search time on HowTo100M is (12.05 + 143.25 = 155.3) ms for text-to-video retrieval and (33.16 + 143.25 = 176.41) ms for video-to-text retrieval,
which is acceptable in practice.

\begin{table}[h]
    \centering
    %\footnotesize
    \scalebox{.85}{
    \begin{tabular}{l|l|llll}
    %\toprule
    %\hline
    Dataset & \# Samples & Visual Embedding & Text Embedding & Similarity Ranking \\
    %\midrule
    \hline
    MSR & 1K & 33.16 ms & 12.05ms & 1.51ms \\
    HowTo100M & 128.94M & 33.16ms & 12.05ms & 143.25ms    \\
    %\bottomrule
    \hline
    \end{tabular}
    }
    \caption{
    Running time analysis for \ModelName during retrieval/inference. 
    Numbers are averaged over 1000 runs.
    }
    \vspace{-2em}
    \label{tab:retrieval_time}
\end{table}

\section{Visualization (Cont'd)}
\begin{figure}[h]
    \centering
        \includegraphics[width=\columnwidth]{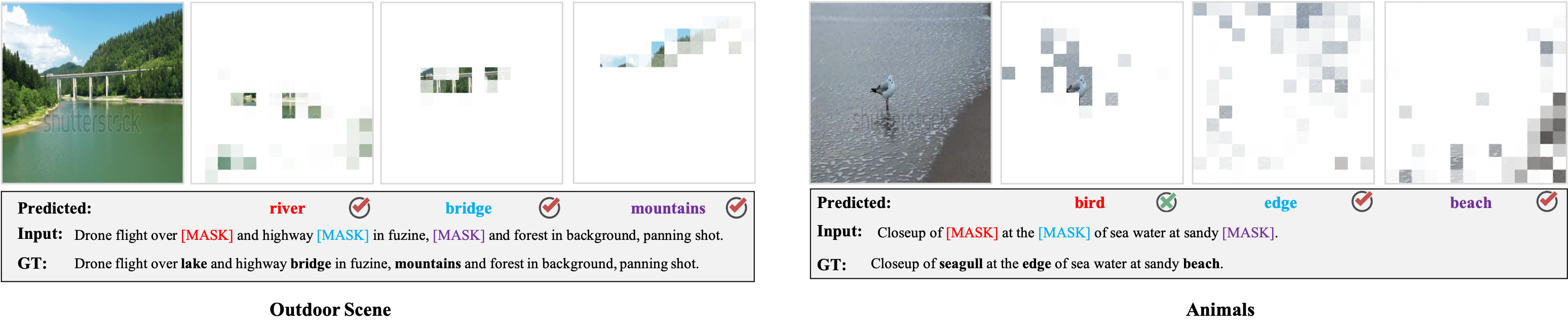}
        \caption{
        Cloze evaluation on \textit{Outdoor Scene} and \textit{Animal} examples.
        }
        \vspace{-2em}
        \label{fig:attn_visualize}
\end{figure}

In addition to the \textit{person-centric} videos, we also visualize some samples about \textit{outdoor scene} and \textit{objects} in Fig. \ref{fig:attn_visualize}.
For find our model can make correct prediction of masked words.

\end{document}